\documentclass[sigconf,nonacm=true]{acmart}

\usepackage{latexsym}
\usepackage{amsmath}
\usepackage{amsthm}
\usepackage{booktabs}
\usepackage{enumitem}
\usepackage{graphicx}
\usepackage{color}
\usepackage{tcolorbox}
\usepackage{tabularx}
\usepackage{multirow}

\newcommand{\etal}{\textit{et al.}}

\begin{document}
\title{Large Language Model Agent for Fake News Detection}

\author{Xinyi Li}
\affiliation{%
\institution{Northwestern University}
\state{Illinois}
\country{United States}
}

\author{Yongfeng Zhang}
\affiliation{%
\institution{Rutgers University}
\state{New Jersey}
\country{United States}}

\author{Edward C. Malthouse}
\affiliation{%
\institution{Northwestern University}
\state{Illinois}
\country{United States}}

\begin{abstract}
In the current digital era, the rapid spread of misinformation on online platforms presents significant challenges to societal well-being, public trust, and democratic processes, influencing critical decision making and public opinion. To address these challenges, there is a growing need for automated fake news detection mechanisms. Pre-trained large language models (LLMs) have demonstrated exceptional capabilities across various natural language processing (NLP) tasks, prompting exploration into their potential for verifying news claims. Instead of employing LLMs in a non-agentic way, where LLMs generate responses based on direct prompts in a single shot, our work introduces FactAgent, an agentic approach of utilizing LLMs for fake news detection. FactAgent enables LLMs to emulate human expert behavior in verifying news claims without any model training, following a structured workflow. This workflow breaks down the complex task of news veracity checking into multiple sub-steps, where LLMs complete simple tasks using their internal knowledge or external tools. At the final step of the workflow, LLMs integrate all findings throughout the workflow to determine the news claim's veracity. Compared to manual human verification, FactAgent offers enhanced efficiency. Experimental studies demonstrate the effectiveness of FactAgent in verifying claims without the need for any training process. Moreover, FactAgent provides transparent explanations at each step of the workflow and during final decision-making, offering insights into the reasoning process of fake news detection for end users. FactAgent is highly adaptable, allowing for straightforward updates to its tools that LLMs can leverage within the workflow, as well as updates to the workflow itself using domain knowledge. This adaptability enables FactAgent's application to news verification across various domains. 
\end{abstract}

\maketitle

\section{Introduction}
The pervasive nature of social media and online platforms in the modern digital era has exacerbated the spread of fake news, characterized by false or misleading information disguised as credible news. The proliferation of fake news poses critical challenges to societal well-being, public trust, and democratic processes \cite{allcott2017social}, with the potential to incite fear, sway public opinion, and influence critical decision-making \cite{naeem2020covid}. To mitigate the ramifications of fake news dissemination, it is imperative to detect fake news, especially in its early stages before it spreads widely on social platforms.

While fact-checking sites such as PolitiFact, and Snopes employ professionals for manual fact-checking,\footnote{https://www.politifact.com, https://www.snopes.com}  the rapid pace of misinformation in the digital age makes laborious manual efforts time-consuming and unscalable \cite{graves2019fact}. Automated solutions are therefore essential, and in recent years, deep neural network-based models for fact-checking have been developed \cite{popat2018declare, liao2023muser}. Detecting fake news is a multifaceted challenge that entails evaluating aspects like authenticity, author intention, and writing style. Various viewpoints can be taken, such as a knowledge-based approach that compares textual information from news articles against a factual knowledge graph, a style-based approach that examines differences in writing style between fake and real news, and a credibility perspective that examines relationships between news articles and entities like publishers \cite{zhou2019fake}. Additionally, propagation-based methods leverage information provided in news dissemination \cite{zhou2019fake}. Existing supervised learning approaches for fake news detection have demonstrated effectiveness in identifying misinformation. However, these models often require human-annotated data for training. This requirement can pose challenges as annotated datasets may not always be readily available or could be costly to collect in practice.

LLMs have demonstrated impressive performance in various NLP tasks \cite{brown2020language, qin2023chatgpt}, motivating us to explore their potential in fake news detection. The fact-checking process for professionals often involves assembling information from multiple, sometimes conflicting sources into a coherent narrative \cite{graves2019fact}, highlighting the importance of verifying details before publication. In this work, we introduce FactAgent, an innovative agentic approach that harnesses LLMs for fake news detection. The distinction between using LLMs in an agentic versus non-agentic way lies in its operational mode: in a non-agentic approach, the LLM responds to prompts or learns in context to generate responses. In contrast, FactAgent integrates LLMs into its decision-making process by breaking down complex problems into manageable sub-steps within a structured workflow, leveraging not only the LLM's internal knowledge but also external tools to complete each component and collaboratively address the overall task. Our primary contributions can be summarized as follows: 
\begin{itemize}
    \item We propose FactAgent, an agentic approach that utilizes LLMs for fact-checking and fake news detection. FactAgent emulates human expert behavior through a structured workflow where LLMs can integrate both internal knowledge and external tools for news verification throughout sub-steps within the workflow. Unlike human experts, FactAgent achieves enhanced efficiency, and unlike supervised models, it operates without the need for annotated data for training. Moreover, FactAgent is highly adaptable, allowing for easy modification for diverse news domains by adjusting the tools in the workflow.
    \item FactAgent is capable of identifying potential fake news early in its dissemination process without relying on social context information. It provides explicit reasoning for the authenticity of news at each step of the workflow, enhancing interpretability and facilitating user understanding.
    \item We conduct experiments on three real-world datasets, demonstrating FactAgent's effectiveness in achieving high performance. We compare FactAgent's performance following an structured expert workflow and an automatically self-designed workflow. Our experiments underscore the critical role of expert workflow design based on domain knowledge for FactAgent.
    
\end{itemize}

\section{Related Work}
\textbf{Fake News Detection} Existing approaches to fake news detection that do not use social context can be categorized into two main groups: content-based and evidence-based. The content-based approaches focus more on the text pattern within the news articles themselves, including writing style and article stance \cite{popat2016credibility}. These approaches can leverage NLP techniques for analysis \cite{przybyla2020capturing} such that LSTM \cite{hochreiter1997long} and BERT \cite{devlin2018bert}. The evidence-based approaches verify news veracity by examining semantic similarity or conflicts in claim-evidence pairs, often retrieving evidence from knowledge graphs or websites. For example, Popat \etal\ \cite{popat2018declare} introduced DeClarE, utilizing BiLSTM and attention network to model claim-evidence semantic relations. Xu \etal\ \cite{xu2022evidence} developed GET, a unified graph-based model for capturing long-distance semantic dependency. Liao \etal\ \cite{liao2023muser} introduced MUSER, a multi-step evidence retrieval enhancement framework for fake news detection. 
With the recent development of LLMs, researchers have explored whether LLMs can effectively detect fake news using internal knowledge. For instance, Wei \etal\ \cite{wei2022chain} explored their potential by using techniques such as zero-shot prompt, zero-shot Chain-of-Thought (CoT) prompting, few-shot and few-shot CoT prompting. They also utilized LLM-generated rationales to enhance the performance of a supervised model like BERT in fake news detection tasks. 

However, the aforementioned approaches still require annotated data for model training, which limits their ability to handle news requiring knowledge not present in the training data. In contrast, our proposed FactAgent eliminates the need for model training by integrating LLM's semantic understanding with external search engine for evidence retrieval. Zhang \etal\ \cite{zhang2023towards} proposed HiSS, a hierarchical step-by-step prompting approach that integrates LLMs to decompose a news claim into subclaims and uses an external search engine to answer queries when LLMs lack confidence. Unlike HiSS, which employs LLMs in a non-agentic manner, our proposed FactAgent utilizes LLMs in an agentic manner, allowing LLMs to rigorously follow a workflow to collect evidence at each sub-step of the process. FactAgent leverages both LLM's internal knowledge and external search engines to examine the veracity of a news claim. 


\textbf{LLM Agent} The development of LLMs has led to the creation of LLM agents with diverse applications across various domains. For instance, Part \etal\ \cite{park2023generative} designed a sandbox environment featuring virtual entities endowed with character descriptions and memory systems to simulate human behaviors. Liang \etal\ \cite{liang2023encouraging} explored a multi-agent debate framework, demonstrating the collaborative problem-solving capabilities of LLM agents. LLM-based approaches offer several benefits, including the ability to provide rationales based on emotions, content, textual descriptions, commonsense, and factual information. Leveraging these advantages, FactAgent is designed to interpret diverse clues and real-world contexts for fake news detection. Unlike existing approaches that allow LLMs to autonomously design their own problem-solving plans \cite{ge2024openagi}, FactAgent enables LLMs to adhere to a structured workflow, emulating human fact-checkers to complete the fake news detection task using the LLM's internal knowledge and external tools.

\section{Methodology} \label{sec:methodology}
While FactAgent is designed to utilize LLMs in an agentic manner, emulating human expert behavior for fact-checking by decomposing tasks into multiple sub-steps within a workflow and collecting evidence from various perspectives using both LLM's internal knowledge and external tools, our primary methodology or focus is to enable LLMs to follow a structured expert workflow designed using domain knowledge in such an agentic manner.

Given the multifaceted nature of fake news detection and the need for a nuanced understanding of diverse clues and real-world context, such as writing style and common sense \cite{zhou2019fake}, coupled with the strong textual understanding capabilities of LLMs and the potential for hallucinations \cite{li2023helma}, we categorize the tools designed for the LLMs to use within the structured expert workflow into two groups: one that utilizes only the LLM's internal knowledge (i.e.\ Phrase, Language, Commonsense, Standing tools) and another that integrates external knowledge (i.e.\ URL and Search tools). Each tool is tailored with specific assumptions about fake news, as described below.

\textbf{\textit{Phrase\_tool}: }This tool is tailored to scrutinize news claims by examining the presence of sensational teasers, provocative or emotionally charged language, or exaggerated claims. It operates under the assumption that fake news often employs these tactics to attract attention from readers. 

\textbf{\textit{Language\_tool}: }This tool is designed to identify grammar errors, wording errors, misuse of quotation marks, or words in all caps within news claims. It assumes that fake news often includes such errors to overemphasize credibility or attract readers.

\textbf{\textit{Commonsense\_tool}: }This tool utilizes LLM's internal knowledge to assess the reasonableness of news claims and to identify any contradictions with common sense. It operates under the assumption that fake news may resemble gossip rather than factual reporting and could contain elements that contradict common knowledge.

\textbf{\textit{Standing\_tool}: }This tool is specifically crafted for news that is relevant to politics and aims to detect whether the news promotes a particular viewpoint rather than presenting objective facts. It operates under the assumption that fake political news often reinforces existing beliefs or biases held by target audiences. Additionally, it may contribute to polarization by portraying political opponents in a negative light or demonizing certain groups. 

\textbf{\textit{Search\_tool}: }This tool utilizes the SerpApi to search for any conflicting information reported by other media resources.\footnote{https://serpapi.com} It assumes that fake news often contains unconfirmed information with little evidence to support the claims being made. Leveraging the Serpapi API can also mitigate the hallucination issue of LLM by using external knowledge to cross-reference and verify the news claim.

\textbf{\textit{URL\_tool}: }This tool integrates LLM internal and external knowledge to assess if the news claim originates from a domain URL that lacks credibility. It first utilizes LLM internal knowledge to gain an overview of a domain URL. Subsequently, it leverages external knowledge such as past experiences stored in a database containing URLs verified for real and fake news to augment the understanding of the domain URL. The assumption underlying this tool is that fake news often originates from domains that are not credible. The external knowledge database can be updated whenever a news article is verified, ensuring its timely accuracy and reliability.

\begin{figure*}[ht]
\centering
\includegraphics[width=\textwidth]{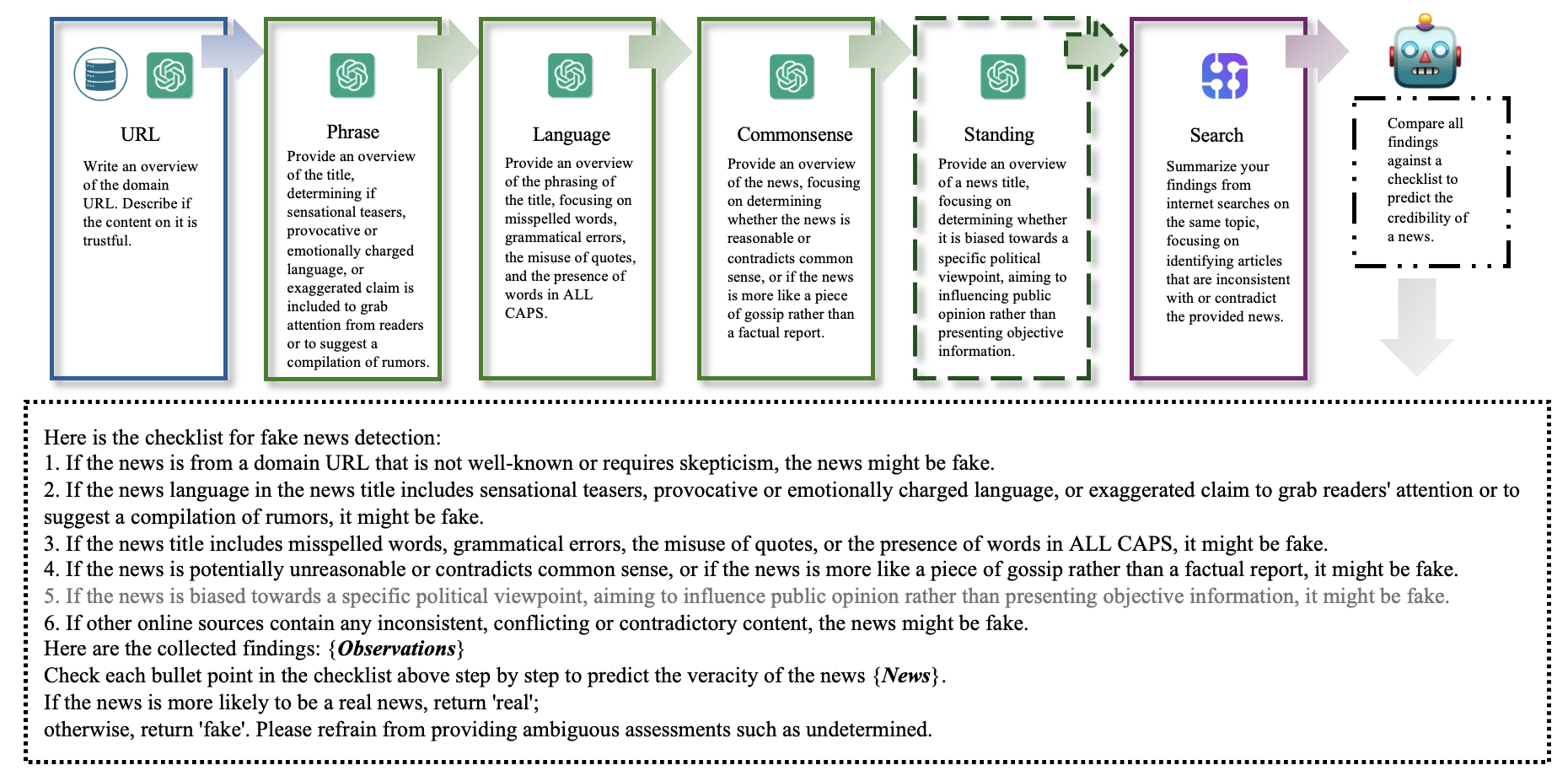}
\caption{
The structured expert workflow for fake news detection is depicted in this diagram. The "Standing\_tool" is highlighted with a dashed frame, and the fifth bullet point is shaded grey to indicate that the "Standing\_tool" and its corresponding checklist item are skipped if the news is not relevant to politics. The \textbf{\textit{Observations}} section comprises a list of observations collected from each tool sequentially. The \textbf{\textit{News}} is represented using its title, domain URL, and publish date, formatted as \textit{`Title: Riverdale Set to Recast a Major Character Ahead of Season 2, Domain URL: tvline.com, Publish Date: 04/25/2017'}. If the domain URL and publish date are unavailable, only the title information is used.} 
\label{fig:workflow}
\end{figure*}

Figure~\ref{fig:workflow} depicts the structured expert workflow, which utilizes the aforementioned tools step by step to gather evidence for verifying the news claim. Upon receiving a news claim,
FactAgent first enables the LLM to utilize its contextual capability to determine if the article concerns politics. If so then the news claim will be analyzed using all provided tools within the structured expert workflow; otherwise the Standing\_tool will be skipped. In the final step, all evidence is collected and compared against a expert checklist to summarize and predict the veracity of the news claim.

\section{Experiments and Results}
We conduct experimental studies to answer the following research questions: 
\begin{itemize}
\item \textbf{RQ1:} How does FactAgent with a structured expert workflow perform compared to other fake news detection baselines?
\item \textbf{RQ2:} How does the domain knowledge influences the FactAgent's performance in fake news detection?
\item \textbf{RQ3:} What is the significance of the external search engine on FactAgent's performance?
\item \textbf{RQ4:} How does the decision-making strategy influence the performance of FactAgent?
\end{itemize}

\subsection{Experimental Setup}
\textbf{Dataset} We evaluate FactAgent's performance using a structured expert workflow with three English-language datasets, Snopes \cite{popat2017truth}, PolitiFact and GossipCop \cite{shu2020fakenewsnet}. PolitiFact and GossipCop are chosen because they not only provide news titles but also source URLs for each news claim. Given that one tool in the expert workflow searches for relevant articles online and identifies conflicting reports, we ensure that the original source URLs of the sampled testing data contain their publication dates. This information is utilized to set constraints for the SerpAPI to avoid the online data leakage problem, wherein events occurring at the current time step are mistakenly included in the search results even though they had not occurred yet when the article was published. For the Snopes dataset, where source URLs are unavailable for news claims, we showcase the flexibility of FactAgent by adjusting the tools used in the expert workflow based on the available information. We randomly select 100 news articles from each dataset for evaluation, ensuring that the ratio of real news to fake news in the testing data is less than 1:2.

\textbf{Baselines} 
To test the effectiveness of FactAgent using a structured expert workflow, we compare it with the following methods: 

\begin{itemize}
\item LSTM \cite{hochreiter1997long}: Applies LSTM to encode textual information from news claims.
\item TextCNN \cite{kim2014convolutional}: Utilizes CNN to capture local patterns and representations of news claims.
\item BERT \cite{devlin2018bert}: Implements a transformer architecture to capture contextual relationships and understand news claims.
\item HiSS \cite{zhang2023towards}: Introduces a hierarchical prompting method directing LLMs to break down claims into sub-claims and verifies them via question-answering steps, leveraging a search engine for external information.
\item Zero-shot Standard Prompt: Utilizes a prompt containing only the task description and the provided news claim.
\item Zero-shot CoT: Applies the CoT \cite{wei2022chain} prompting approach for zero-shot inference.
\item Zero-shot (Tool): Leverages individually designed tools mentioned in Section~\ref{sec:methodology}, excluding the Standing\_tool since not all news concern politics. 
\end{itemize}

\textbf{Implementation Details} We employ the LangChain framework and \textit{gpt-3.5-turbo} model as the underlying LLM for all tools employed in the analysis engine.\footnote{https://www.langchain.com} The temperature parameter is set to 0 to ensure the reproducibility. Each article is represented using its title, domain URL, and publication date for PolitiFact and GossipCop dataset. For the Snopes dataset only the title information is used due to the unavailability of URLs. Other baselines in our comparison used only the news title for analysis. The statistics of the baseline training data are summarized in Table~\ref{tab:statistics}. 

For the baseline HiSS \cite{zhang2023towards}, we directly utilize the same prompt provided by the researchers 
with the modification that `pants-fire', `barely-true', and `false' labels are grouped together as `false', while  `half-true', `mostly-true', and `true' labels are grouped as `true' following Rashkin \etal\ \cite{rashkin2017truth}.

To ensure an equitable comparison of LLMs using individually designed tools and FactAgent following the designed expert workflow, we retain the analysis results at each step from each tool throughout the workflow process and then apply the same bullet point for each tool in Figure~\ref{fig:workflow} methodology to derive the final prediction veracity results for LLM using each tool in a zero-shot manner.

To assess performance, we employ various metrics including accuracy, F1 score, as well as F1 scores specifically for real and fake news classifications to get a comprehensive assessment of the model's performance in fake news detection.

\begin{table}[h]
    \centering
    \caption{Statistics of datasets for supervised baselines training.}
    \begin{tabularx}{\columnwidth}{lXXX} 
        \hline
         & \textbf{GossipCop} & \textbf{PolitiFact} & \textbf{Snopes} \\
        \hline
        \#Real News & 3586 & 456 &1050 \\
        \#Fake News &  2884 & 327 &1500 \\
        \hline
    \end{tabularx}
    \label{tab:statistics}
\end{table}

\section{Experimental Results}

\subsection{Fake News Detection Performance (RQ1) }

The performance comparisons of different models are summarized in Table~\ref{tab:result}. Our observations reveal that FactAgent, following the expert workflow, achieves superior performance compared to other baselines on all datasets.

Unlike supervised baselines that implicitly learn contextual patterns or writing styles of fake news from labeled data, the designed tools explicitly leverage the LLM's internal knowledge and contextual understanding capabilities to assess the specific existence of common phrasing or language styles indicative of fake news claims. FactAgent also benefits from external tools that extend the LLMs capabilities beyond what supervised baselines can achieve. For example, FactAgent enables LLMs to search for related news claims online and detect conflicting reports, leveraging external sources to enhance their veracity assessment. Additionally, FactAgent allows LLMs to check the credibility of domain URLs using both internal knowledge of domain URLs and external databases containing recent domain URLs associated with verified fake news. These capabilities enable FactAgent with a structured expert workflow to have enhanced performances without requiring model training and hyper-parameter tuning processes typically associated with supervised learning models. 

Comparing the performances of LLMs using different prompting techniques (i.e., standard prompt, CoT prompt), or making decisions based on findings from each designed tool, we observe that LLMs utilizing CoT prompting do not consistently outperform standard prompting techniques. This observation is consistent with previous research \cite{zhang2023towards}, which also highlighted similar conclusions. After analyzing errors associated with the CoT prompting approach, we also observe the omission of necessary thought, echoing the finding of Zhang \etal\ \cite{zhang2023towards}. The inferior performances of standard prompting and CoT prompting compared to LLMs making decisions based on commonsense and phrase observations across all three datasets have underscore the importance of explicitly guiding the LLM reasoning process from a specific perspective to effectively leverage its textual understanding capability. Furthermore, the superior performance of the FactAgent with an expert workflow compared to the LLM using individual tools highlights the importance of examining a news claim from various perspectives.

Our approach differs from HiSS, which utilizes an LLM and external search engines primarily as a prompting technique. In contrast, FactAgent decomposes the fake news detection problem into simple tasks, with each task relying on LLMs to provide answers. The superiority of FactAgent over HiSS is attributed to the rigorous utilization of external tools integrated into the structured expert workflow, along with the examination of the LLM's internal commonsense. In contrast, HiSS relies on external search engines only when the LLM itself lack confidence to answer specific questions, potentially limiting its scope and depth of analysis compared to FactAgent. 

Overall, the superior performance of FactAgent over baselines demonstrates the advantage of utilizing LLMs in an agentic way to emulate human expert behaviors, rigorously examining a news claim from multiple perspectives, and integrating an external search process following an expert workflow to verify its veracity. 

\begin{table*}
\caption{Performance comparison among different models for PolitiFact, GossipCop, and Snopes. The superior outcomes are indicated in bold, with statistical significance indicated by a p-value < 0.05.}
\begin{tabularx}{\linewidth}{lXXXXXXXXXXXX}
\hline
\multirow{2}{*}{Model} & \multicolumn{4}{c}{PolitiFact} & \multicolumn{4}{c}{GossipCop} & \multicolumn{4}{c}{Snopes} \\ \cmidrule(lr){2-5} \cmidrule(lr){6-9} \cmidrule(lr){10-13} 
    &   F1 & Acc. & F1\_real & F1\_fake & F1 & Acc. & F1\_real & F1\_fake& F1 & Acc. & F1\_real & F1\_fake \\  \hline
LSTM & 0.79 & 0.79 & 0.79 & 0.79  & 0.77 & 0.77 & 0.76 & 0.77 & 0.65& 0.66 & 0.64 & 0.67\\

TextCNN & 0.80 & 0.80 & 0.79 & 0.82 & 0.79& 0.79 & 0.78 & 0.79 & 0.62 & 0.64& 0.55 & 0.69\\

BERT & 0.85 & 0.85 & 0.85 & 0.85 & 0.79& 0.79 & 0.78 & 0.80 & 0.63 & 0.63& 0.59& 0.67\\

HiSS & 0.62 & 0.62 & 0.58 & 0.65 & 0.66 & 0.66 & 0.69 & 0.63 & 0.58 & 0.60 & 0.47 & 0.68\\
Zero-shot Standard Prompt & 0.73 & 0.73 & 0.70 & 0.76 & 0.61 & 0.61 & 0.65 & 0.56 & 0.61 & 0.62& 0.55& 0.67 \\

Zero-shot CoT Prompt & 0.64 & 0.63 & 0.56 & 0.69 & 0.64 & 0.64 & 0.63 & 0.65 & 0.59 & 0.63& 0.46& 0.72\\

Zero-shot Language Tool & 0.73 & 0.72 & 0.77 & 0.67 & 0.53 & 0.48 & 0.64 & 0.32 & 0.60 & 0.62 & 0.53 & 0.68\\

Zero-shot Phrase Tool &  0.83 & 0.83 & 0.85 & 0.80 & 0.69 & 0.69 & 0.71 & 0.67 & 0.66 & 0.68 & 0.75 & 0.57\\

Zero-shot URL Tool &  0.81 & 0.81 & 0.83 & 0.79 & 0.63 & 0.63 & 0.64 & 0.62 & ------ & ------ & ------ & ------\\

Zero-shot Search Tool &  0.78 & 0.78 & 0.77 & 0.79 & 0.66 & 0.66 & 0.67 & 0.65 &  0.72 & 0.73 & 0.67 & 0.77\\

Zero-shot Commonsense Tool &  0.80 & 0.80 & 0.80 & 0.80 & 0.76 & 0.75 & 0.71 & 0.79 & 0.66 & 0.66 & 0.61 & 0.70  \\
\hline 
FactAgent with Expert Workflow &  \textbf{0.88} & \textbf{0.88} & \textbf{0.89}& \textbf{0.88} & \textbf{0.83} & \textbf{0.83} & \textbf{0.83} & \textbf{0.83} & \textbf{0.75} & \textbf{0.75} & \textbf{0.75} & \textbf{0.75}\\
\hline 
\end{tabularx}
\label{tab:result}
\end{table*}

\subsection{Importance of Domain Knowledge (RQ2)}
This subsection conducts two experiments to evaluate the importance of domain knowledge in creating the expert workflow for FactAgent to verify the veracity of news claims. 

\begin{figure}[h]
    \centering
    \includegraphics[width=\columnwidth]{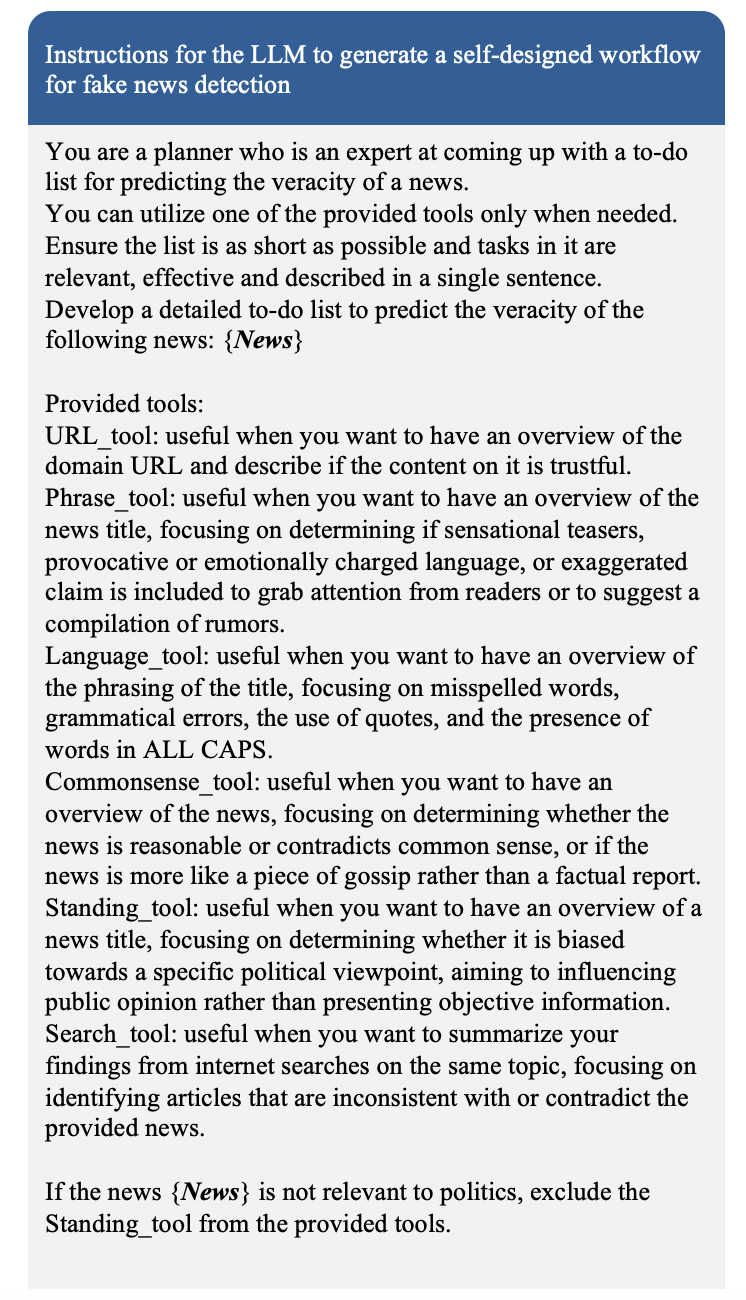}
    \caption{Instructions for the LLM to automatically generate a self-designed workflow for fake news detection. The \textbf{\textit{News}} is represented using its title, domain URL, and publish date if available.}
    \label{fig:plan_prompt}
\end{figure}

\textbf{Expert Workflow vs. Automatically Self-designed Workflow} Figure~\ref{fig:plan_prompt} depicts the instructions for the LLM to automatically generate a self-designed workflow to verify a news claim using the provided tools. The agent then executes the chosen tools step by step to collect findings related to the news claim. This process is similar to the workflow shown in Figure~\ref{fig:workflow}. At the final step, the LLM compares its findings with the checklist items corresponding to the selected tools to determine the veracity of the news claim.

\begin{figure}[h]
\centering
\includegraphics[width=\columnwidth]{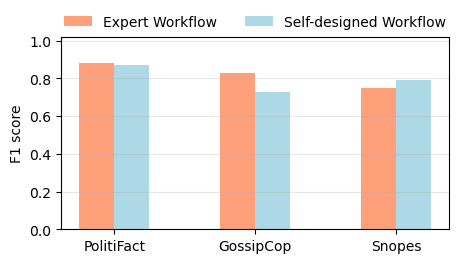}
\caption{Performance comparison of FactAgent following an automatically self-designed workflow and an expert workflow.}
\label{fig:RQ2}
\end{figure}

\begin{figure}[h]
\centering
\includegraphics[width=\columnwidth]{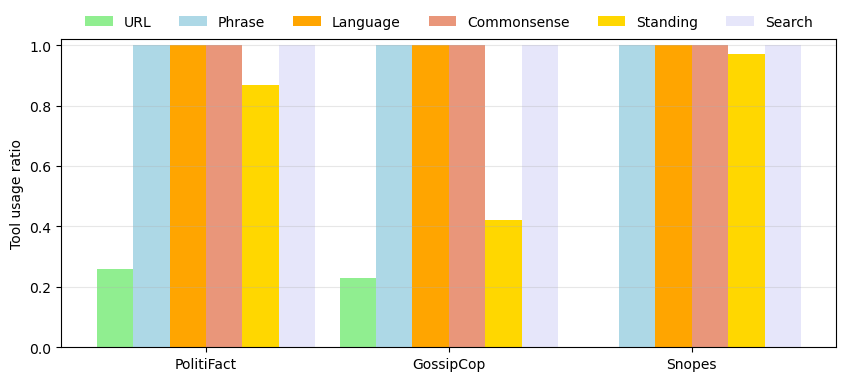}
\caption{The frequency of each tool's usage among the testing samples when the LLM creates a self-designed workflow to evaluate news veracity.}
\label{fig:ratio_tool}
\end{figure}

Figure~\ref{fig:RQ2} compares the performance of FactAgent with an expert workflow against that of an automatically self-designed workflow. It is observed that allowing FactAgent with an LLM to automatically design its own workflow for fake news detection results in inferior performance compared to instructing the agent to adhere to an expert workflow for the GossipCop dataset. The performance is similar for PolitiFact dataset, and shows slightly better performance for Snopes dataset. We further analyze the ratio of tool usage among testing samples for each dataset when the LLM designs its own workflow in Figure~\ref{fig:ratio_tool}. From Figure~\ref{fig:ratio_tool}, we observe that when the LLM designs the workflow using provided tools, it tends to prefer tools that focus more on news textual content while neglecting other important factors such as domain URL, despite the availability of the domain URL in the news description.

\begin{figure}[h]
\centering
\includegraphics[width=\columnwidth]{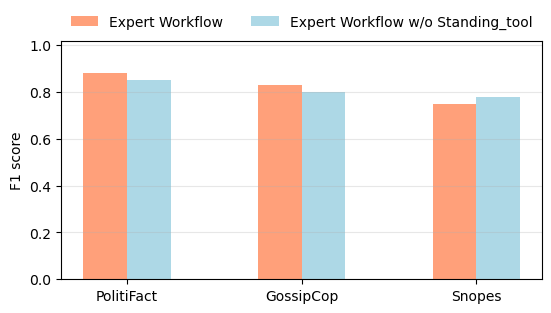}
\caption{Performance comparison of FactAgent with and without utilizing the Standing\_tool within the expert workflow.}
\label{fig:standing}
\end{figure}

\textbf{Integration of Standing\_tool for Political News}
The Standing\_tool is specifically designed to check news claims that the LLM identifies as relevant to politics. In Figure~\ref{fig:standing}, we conduct an experiment where we remove the Standing\_tool from the expert workflow shown in Figure~\ref{fig:workflow}. As a result, all news articles, regardless of their relevance to politics, are processed using the remaining general tools (excluding the URL\_tool for the Snopes dataset). For the PolitiFact and GossipCop datasets, all remaining tools are applied. Figure~\ref{fig:standing} shows the results of this experiment, which indicate a decline in performance for the PolitiFact and GossipCop datasets, and slightly better performance for the Snopes dataset.

Combining the above observations from Figure~\ref{fig:RQ2}, Figure~\ref{fig:ratio_tool} and Figure~\ref{fig:standing} in the context of PolitiFact data, we deduce that the Standing\_tool, which analyzes political views, holds greater importance than the URL\_tool. Conversely, for GossipCop data, we infer that overusing the Standing\_tool and underutilizing the URL\_tool produces worse results compared to scenarios where the URL\_tool is used but the Standing\_tool is omitted. Given that GossipCop data primarily focuses on entertainment, celebrity gossip, and rumors rather than politics, the emphasis on political bias detection with the Standing\_tool in LLM self-designed workflow may not be as relevant for this dataset. 

For the Snopes dataset, both using and not using the Standing\_tool show improvements in performance. This suggests that identifying whether news is relevant to politics is less critical for Snopes dataset. The external Search\_tool appears to play a significant role in achieving good performance from Table~\ref{tab:result}. Combining the Search\_tool with textual pattern analysis tools seems sufficient for the LLM to summarize the overall findings when combined with external search results.

These observations underscore the importance of structuring an expert workflow that incorporates domain knowledge to design appropriate tools specific to the dataset's domain, rather than allowing the LLM to automatically design its own workflow for FactAgent. This approach ensures that the designed workflow is more tailored and suitable for the characteristics of the news domain being analyzed. One of the key benefits of FactAgent is the flexibility and ease of adding, deleting, or modifying certain tools within the workflow based on domain-specific requirements. This flexibility allows adaptations for different datasets and contexts, enhancing the effectiveness and applicability of the FactAgent in various scenarios.

\subsection{Importance of External Search Engine (RQ3)}

Table~\ref{tab:result} shows that solely relying on external searches to detect conflicting reports as a means of scrutinizing news claims does not yield optimal performance, particularly for the GossipCop dataset. This outcome could be attributed to the fact that while online searches can offer evidence, the same rumor or misinformation may also be reported across multiple online resources, leading to a dilution of credibility rather than a clear verification. 

\begin{figure}[h]
\centering
\includegraphics[width=\columnwidth]{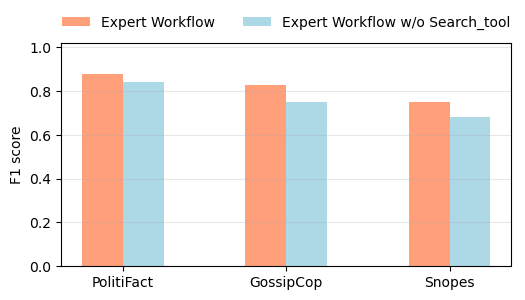}
\caption{Performance comparison of FactAgent with and without utilizing the Search\_tool within the expert workflow.}
\label{fig:RQ3}
\end{figure}

Despite the suboptimal performance of relying solely on external searches, this subsection evaluates the impact of incorporating the external search tool within the expert workflow for FactAgent. To test the importance of the Search\_tool, we exclude it from the expert workflow, leaving the LLM to assess veracity based solely on the available tools, which primarily utilize the LLM's internal knowledge. Figure~\ref{fig:RQ3} indicates that performance deteriorates without the Search\_tool, suggesting that relying solely on the LLM's internal knowledge is insufficient for effective fake news detection.

\subsection{Ablation Study on Decision Making Strategy (RQ4)}

This subsection explores the final decision-making strategy employed by the LLM at the final step of FactAgent with an expert workflow. Currently, the LLM autonomously compares the collected information from various tools and makes the final decision without predefined decision rules, such as relying on each tool's prediction and using a majority vote strategy.

\begin{figure}[h]
\centering
\includegraphics[width=\columnwidth]{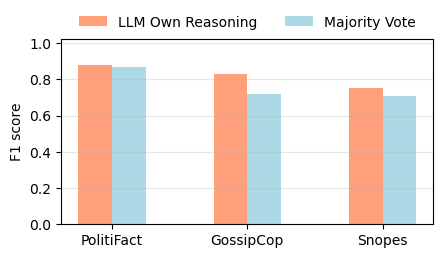}
\caption{Performance comparison of different decision-making strategies.}
\label{fig:RQ5}
\end{figure}

To investigate the impact of using a majority vote strategy during the final verification process, we manually apply a majority vote to all tools' decisions used by the LLM. Figure~\ref{fig:RQ5} shows that using majority voting results in inferior performance across all three datasets compared to instructing the LLM to compare with a checklist for the final prediction. This performance discrepancy highlights that when the LLM compares observations with the checklist at the final step of FactAgent, it does not blindly rely on a majority vote. Instead, it suggests that giving the LLM the flexibility to summarize the overall prediction based on its reasoning and insights may yield better outcomes than imposing rigid decision rules like a majority vote.

\section{Case Study}
\begin{figure}[h]
    \centering
\includegraphics[width=\columnwidth]{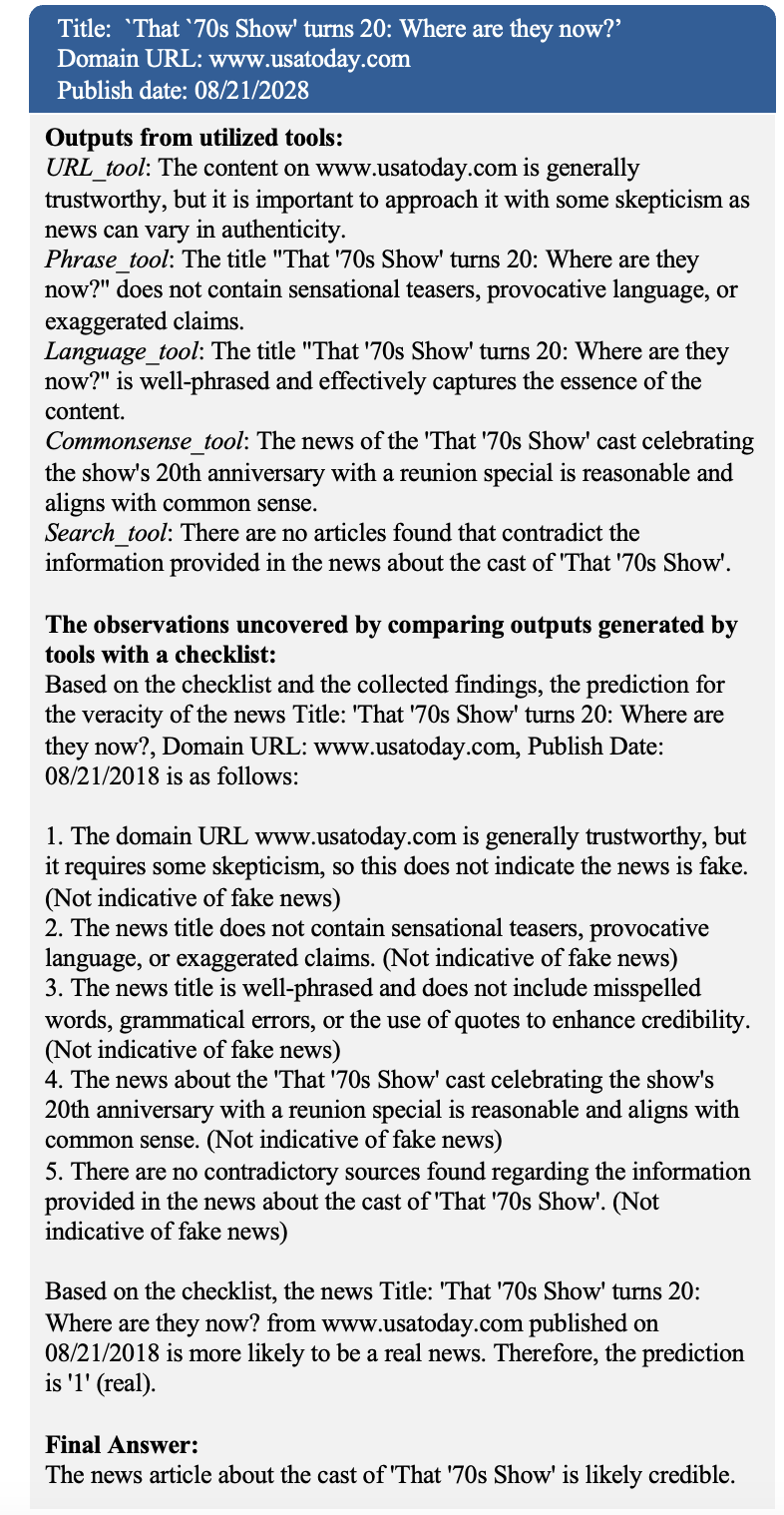}
\caption{A case demonstrating the FactAgent's reasoning process at each step of the expert workflow, leading to the final conclusion. The LLM does not identify this news as relevant to politics; therefore, the Standing\_tool is not used.}
\label{fig:case}
\end{figure}

In the context of fake news detection, interpretation is critical for providing clarity and understanding to end users. 
Figure~\ref{fig:case} illustrates the reasoning process of the LLM at each step of the workflow within FactAgent. The LLM outputs explicit observations from each tool in natural language and provides reasoning at the final step by comparing each observation with the corresponding checklist item to reach a conclusion. This stands in contrast to conventional supervised models that may lack explicit transparency in their decision pathways. The checking and reasoning process can also inform end users about what aspects to scrutinize to avoid falling for fake news.

\section{Conclusion}
Motivated by the rich internal knowledge embedded within LLMs, our work proposes FactAgent, which uses LLMs in an agentic manner to emulate human expert behavior in assessing news claims via a structured workflow. Our results demonstrate that FactAgent with an expert workflow outperforms supervised learning models, standard prompting, CoT prompting techniques, and single-aspect analyses. By rigorously integrating the external search and LLM's commonsense within the workflow, FactAgent also outperforms HiSS, which also utilizes LLM and external search engine for fact checking. Furthermore, our experiments underscore the importance of leveraging domain expert knowledge to design FactAgent's workflow and highlight the flexibility of modifying the workflow and the final decision-making strategy.

The benefits of our approach over existing ones are many. Unlike supervised learning models that require annotated data for training and time-consuming hyper-parameter tuning, our method does not require any training or tuning. This makes our approach highly efficient and accessible, eliminating the need for extensive manually labeled datasets. Moreover, our experiments demonstrate that FactAgent's performance relies on expert domain knowledge used to design the workflow. While the expert workflow designed in this work may not be optimal, a key advantage of FactAgent is its flexibility in incorporating new tools into the workflow. When experts detect new indicators of fake news in a specific domain, they can easily integrate this knowledge into a tool and seamlessly merge it with the existing workflow. FactAgent can then instruct the LLM to simulate their fact-checking process. This benefits have significant implications, such as using FactAgent with a carefully designed workflow to assist humans in data annotation for training supervised models for fake news detection. Additionally, the FactAgent provides explicit reasoning for each step in the workflow, enhancing the interpretability of the fact-checking process in natural language.

Some limitations in our work suggest potential directions for future research. For instance, our approach currently relies on news titles and domain URLs if available, but considering a social context such as retweet relationships could be important for detecting fake news during the dissemination process \cite{zubiaga2018detection}. Additionally, incorporating a multi-modal approach by analyzing web design elements may also enhance fake news detection capabilities. Moreover, some news titles alone may not directly determine the veracity of the news claim. It could be beneficial to integrate the full content of news articles and assess whether the title accurately reflects the content or if it is misleading to attract user engagement (i.e.\ clickbait). Lastly, exploring the integration of expert decision-making strategies to improve performance represents another potential avenue for future investigation.

\bibliographystyle{ACM-Reference-Format}
\bibliography{ref}

\end{document}